\documentclass[sn-mathphys,Numbered]{sn-jnl}

\usepackage{graphicx}%
\usepackage{caption}
\usepackage{subcaption}
\usepackage{multirow}%
\usepackage{amsmath,amssymb,amsfonts}%
\usepackage{amsthm}%
\usepackage{amssymb}
\usepackage{adjustbox}
\usepackage{amsfonts}
\usepackage{mathrsfs}%
\usepackage[title]{appendix}%
\usepackage{xcolor}%
\usepackage{textcomp}%
\usepackage{manyfoot}%
\usepackage{booktabs}%
\usepackage{algorithm}%
\usepackage{algorithmicx}%
\usepackage{algpseudocode}%
\usepackage{listings}%
\usepackage{hyperref}
\usepackage[T1]{fontenc} 
\usepackage{booktabs}
\usepackage[inkscapearea=page]{svg}
\DeclareMathOperator*{\argmin}{arg\,min}

\begin{document}

\title[Medical images interpretability and explaonability]{Visual Interpretable and Explainable Deep Learning Models for Brain Tumor MRI and COVID-19 Chest X-ray Images}


\author*[1]{\fnm{Yusuf} \sur{Brima}}\email{ybrima@uos.de}

\author*[2]{\fnm{Marcellin} \sur{Atemkeng}}\email{m.atemkeng@ru.ac.za}

\affil*[1]{\orgdiv{Computer Vision}, \orgname{Institute of Cognitive Science, Osnabrück University}, \orgaddress{\street{}, \city{Osnabrueck}, \postcode{D-49076}, \state{Lower Saxony}, \country{Germany}}}

\affil[2]{\orgdiv{Rhodes AI Research Group}, \orgname{Department of Mathematics, Rhodes University}, \orgaddress{\street{}, \city{Grahamstown}, \postcode{6140}, \state{Eastern Cape}, \country{South Africa}}}

\abstract{
 Deep learning shows promise for medical image analysis but lacks interpretability, hindering adoption in healthcare. Attribution techniques that explain model reasoning may increase trust in deep learning among clinical stakeholders. This paper aimed to evaluate attribution methods for illuminating how deep neural networks analyze medical images. Using adaptive path-based gradient integration, we attributed predictions from brain tumor MRI and COVID-19 chest X-ray datasets made by recent deep convolutional neural network models. The technique highlighted possible biomarkers, exposed model biases, and offered insights into the links between input and prediction. Our analysis demonstrates the method's ability to elucidate model reasoning on these datasets. The resulting attributions show promise for improving deep learning transparency for domain experts by revealing the rationale behind predictions. This study advances model interpretability to increase trust in deep learning among healthcare stakeholders. 
}




\maketitle

\section{Introduction}\label{intro}
Recent advances in compute and deep neural architectures~\cite{rumelhart1986learning,he2016deep, chollet2017xception,krizhevsky2012imagenet,simonyan2014very} have enabled rapid progress in automated medical image analysis. Medical imaging techniques like Computed Tomography (CT), Magnetic Resonance Imaging (MRI), Functional Magnetic Resonance Imaging (fMRI), Positron Emission Tomography (PET), Mammography, Ultrasound, and X-ray are traditionally interpreted by radiologists and physicians for timely disease detection and diagnosis~\cite{litjens2017survey}. However, the healthcare field's high demand for skilled labor can lead to fatigue, necessitating computer-aided diagnostic tools. The maturation of deep learning is thus accelerating the adoption of computer-assisted tools to aid experts and reduce manual analysis.

Deep learning shows particular promise for democratizing healthcare globally by reducing prohibitive costs of expertise~\cite{murtaza2020deep}. However, successful clinical adoption depends on assured trust in model robustness and interpretability, which is crucial in safety-critical healthcare~\cite{reyes2020interpretability}. Despite the inherent complexity of deep learning models, we present techniques to illuminate their inference mechanisms. By this, we refer to how a deep model takes an input (e.g., a medical image) and produces an output prediction (e.g. a disease classification).

Using adaptive path-based integrated gradients, we systematically studied model predictions on brain tumor MRI~\cite{Cheng2017} and COVID-19 chest X-rays~\cite{9144185} medical images. Attribution maps highlighted salient input features corresponding to model predictions. These techniques can build understanding, trust, and verification by experts to enable the adoption of computer-aided diagnostics.

In this work, we aim to evaluate attribution methods on convolutional neural networks (CNNs) analyzing medical images (Section~\ref{methods}). Experiments assess technique effectiveness across models and modalities (Section~\ref{results}). Our results demonstrate the ability of these attribution methods to provide insights into input-prediction relationships, reveal potential biomarkers, and uncover model biases.

This work makes key contributions through a comprehensive evaluation of adaptive gradient-based attribution methods across diverse CNNs and medical imaging datasets. Visualizations demonstrate clear technique differences and reveal relationships to model structure. 

The paper is organized as follows. Related interpretability approaches are discussed in Section~\ref{literature}. Section~\ref{methods} describes the methodology. Section~\ref{results} presents experimental results on three datasets. Section~\ref{conclusion} concludes and proposes future directions. Together, this work advances model transparency to increase trust in deep learning for medical image analysis.
\section{Related Literature}
\label{literature}
Varied interpretability methods have been recently proposed for medical image analysis tasks. Research in this direction is growing primarily to help build trustworthy artificial intelligence (AI) systems that use a human-in-the-loop approach to complement domain experts. Concept Learning techniques have been used in~\cite{koh2020concept,sabour2017dynamic,shen2019interpretable} to manipulate high-level concepts to train models that can perform multi-stage predictions from high-level clinical concepts which provide input to the final classification task of disease categories. However, these methods have significant annotation costs, and concept-to-task mismatches can lead to considerable information leakage~\cite{SALAHUDDIN2022105111}.

Another class of technique is Case-Based Models (CBMs), where class discriminative disentangled representations and feature mappings are learned. The final classification is performed by measuring the similarity between the input image and the base templates~\cite{bass2020icam,kim2021xprotonet,li2018deep}. But this class of techniques is not susceptible to corruption by noise and compression artifacts. It is also difficult to train models using this paradigm. Counter Factual Explanation is another approach where input medical images are perturbed in pseudo-realistic ways to generate an opposite prediction. They have the problem of generating unrealistic perturbations with respect to the input images which can often be low resolutions as opposed to the original images~\cite{baumgartner2018visual,cohen2021gifsplanation,lenis2020domain,schutte2021using,seah2019chest,selvaraju2017grad,simonyan2013deep,singla2021explaining}. Visualization of the internal network representation of learned features of kernels in CNNs is another technique that is used in model understanding. But this approach has a limitation of difficulty in interpreting feature maps in medical image analysis settings~\cite{bau2017network,natekar2020demystifying}.

An attribution map provides post-hoc explanations whereby regions of the input image are highlighted as indicated saliency method based on the model prediction. In their paper,~\cite{bohle2019layer} proposed layer-wise relevance propagation for explaining deep neural network decisions in MRI-based Alzheimer's disease classification. A deep CNN-based model with Gradient Class Activation Map (Grad-CAM) was trained to classify oral lesions for clinical oral photographic images~\cite{camalan2021convolutional}. In~\cite{kermany2018identifying}, a  similar CNN-based Grad-CAM technique for the classification of Oral Dysplasia is proposed. However, our approach is different from~\cite{bohle2019layer,camalan2021convolutional,kermany2018identifying} as we utilize adaptive path-based integrated gradients techniques to address the problem of noisy saliency masks which hinders former methods~\cite{kapishnikov2021guided}.
\section{Methods}
\label{methods}
We present the CNN models utilized to carry out experiments in this study for the classification tasks. Characterizations of these CNN architectures are expounded, indicating their inductive priors, strengths, and limitations in learning visual representations. We give a detailed description of the adaptive path-based integrated gradient techniques and their direct applications to deep learning-based models in medical image analysis. To achieve this, we have summarized the mathematical notation in Table~\ref{tab:notation} used in this work.
\begin{table}[ht!]
\centering
\caption{\label{tab:paper_notation} A summary of the mathematical notations in this paper.}
\begin{tabular}{ll}
\toprule Notation & Description \\ \addlinespace[5pt] \cmidrule(lr){1-2}
$\mathbb{R}$ & Set of real numbers \\
$\mathbb{R}^{d}$ & Set of $d$-dimensional real-valued vector \\
$\mathbb{R}^{n \times d}$ & Set of $n \times d$ real-valued matrix \\
$\mathbf{x} \in \mathbb{R}^{n \times d \times 1}$  & Set of $n \times d \times 1$ real-valued tensor which is a single channel image input to a neural network\\
$\mathbf{y} \in \mathbb{R}^{|C|}$ & A corresponding one-hot encoded label for an image input $\mathbf{x}$\\
$|C|$ & Cardinality of the set of medical image classes.\\ 
$W_{i}$ & The kernels for the $i$-th layer of a CNN\\
$\mathcal{L}(\cdot)$ & A loss function\\
$f^l(\mathbf{x}^m, \boldsymbol{\theta})$ & Non-linear transformation of input $\mathbf{x}^m$ at layer $l$ parameterized by $\boldsymbol{\theta}$ \\
$\sigma^{l}$ & Activation function at layer $l$\\
$\alpha \in \mathbb{R}_+$ & Non-negative real-valued regularization hyperparameter\\
$|| \cdot ||_2^2$ & The squared $\ell_{2}$ norm\\
$\mathcal{D}_{i}$ and $\mathcal{D}_i^\prime$ & A training and testing samples of task $\mathcal{T}_i$ respectively.  $\mathcal{T}_i$ is sampled from the distribution of task \\
$h(\cdot)$ & A neural network that produces latent representation for each input \\
$A_h$ & An attribution operator that takes a trained model $h$ to produce a saliency map\\
$\hat{\mathbf{x}}$ &  Computed saliency map for a given input image $\mathbf{x}$\\
\bottomrule
\end{tabular}
\label{tab:notation}
\end{table}
\subsection{Background}
We use 9 standard CNN architectures: Visual Geometric Group (VGG16 and VGG19~\cite{simonyan2014very}), Deep Residual Network (ResNet50, ResNet50V2)~\cite{he2016deep}, Densely Connected Convolutional Networks (DenseNet)~\cite{huang2017densely}, Deep Learning with Depthwise Separable Convolutions (Xception)~\cite{chollet2017xception}, Going deeper with convolutions (Inception)~\cite{szegedy2015going}, a hybrid deep Inception and ResNet and EfficientNet: Rethinking model scaling for convolutional neural networks~\cite{tan2019efficientnet} for classifying COVID-19 X-ray images and brain tumors from the T1-weighted MRI slices. The choice of these deep models is explained by the fact that they are modern techniques that are widely used in solving vision tasks and by extension medical image feature extraction for prediction and/or classification.

VGG was first introduced in the ImageNet Large Scale Visual Recognition Challenge (ILSVRC) 2014 challenge~\cite{russakovsky2015imagenet} mainly to evaluate the effect of increasing depth in a deep neural network architecture with very small ($3 \times 3$) convolution kernels. The results  showed that increasing depth from 16 to 19 weight layers is a significant factor in improving the prior-art configurations. Increment in neural architectural depth leads to more expressive models that learn better representations, thus, improving generalizations across training tasks. However, deeper networks are hard to train because of the vanishing gradient problem~\cite{hochreiter1991untersuchungen, bengio1994learning,glorot2010understanding}. In that regard, deep residual learning: ResNet was introduced in~\cite{he2016deep} to facilitate training routines for massively deeper neural networks. Results in~\cite{he2016deep} empirically showed that ResNet converges faster using local search methods such as stochastic gradient descent (SGD) and can achieve higher accuracy from the considerably increased depth of several layers. The primary way the vanishing gradient problem is tackled in this framework is by introducing identity mappings that create shortcut connections to maximally exploit information flow in the network architecture thus solving the vanishing gradient problem. As depth is addressed by the residual network framework, another key concern is how wide can we go and in what variety of kernel sizes.

Thus, a natural solution would be to learn, within computational limits as many factors of variations as possible. This is the main idea introduced in the depth-wise separable layers based on the Inception architecture~\cite{szegedy2015going}. Inspired by the promising performance of both Inception and ResNet, a hybrid model that combines any of the sub-versions (i.e., v1, v2, v3, or v4) of ResNet and Inception has shown satisfactory results when compared to ResNet-only or Inception-only~\cite{szegedy2017inception,alotaibi2020hybrid}. The drawback of the hybrid InceptionResNet is the computational requirements at the training stage.

In contrast to a standard Inception model that performs cross-channel correlations followed by spatial correlations, in the Xception model, spatial convolutions are performed independently~\cite{chollet2017xception}. This consists of a spatial convolution performed independently for each channel of the input followed by a point-wise convolution across channels for dimensionality reduction of the computed features.  In their work~\cite{huang2017densely}, introduced the idea of dense connectivity: DenseNet where each layer is connected to every other layer in a feed-forward fashion in neural networks. Their approach is  an extension of the successes made by ResNets. A DenseNet comprises dense blocks which implement dense connectivity to reduce the computational cost of channel-wise feature concatenation. This architectural design is robust to gradient flow as it provides robust signals for gradient propagation in the layers of a substantially deeper network which results in gainful generalization performance. With a small growth rate, this architectural design is computationally efficient. The EfficientNet~\cite{tan2019efficientnet} introduced a principled study of model scaling considering the impact of depth, width, and resolution on model performance. A new compound scaling method was proposed that uniformly scales all three dimensions of an input image: depth, width, and resolution using a compound coefficient that is derived from a grid search method.

The above architectures as described are known in the context of supervised deep learning for which the optimization uses gradient-based local search methods. The goal of the optimization is to find an optimal fitted function that minimizes the empirical risk; measured from the training samples with a defined loss function $\mathcal{L}$:
\begin{equation}
   \hat{\boldsymbol{\theta}} =\argmin_{\boldsymbol{\theta}}\frac{1}{N}\sum_{m=1}^{N}\mathcal{L}(y^m, f(\mathbf{x}^m; \boldsymbol{\theta})), 
   \label{eq:emrisk}
\end{equation}
where $\boldsymbol{\theta}$ compacts the parameters of the trainable neural network $f(\mathbf{x}^m;\boldsymbol{\theta})$, $N$ the number of training examples, $\mathbf{x}^m$   and associated $y^m$ are  the  features vector and label for  sample $m$ respectively. To prone generalization, a regularization term is imperatively added
\begin{equation}
    \hat{\boldsymbol{\theta}}=\argmin_{\boldsymbol{\theta}}\frac{1}{N}\sum_{m=1}^{N}\mathcal{L}(y^m, f(\mathbf{x}^m; \boldsymbol{\theta})) +\alpha||\boldsymbol{\theta}||_2^2
    \label{eq:emriskg}
\end{equation}
in the $L_2$ norm regime with $\alpha$ the learning rate. In order words for $f (\mathbf{x}^m;\boldsymbol{\theta}) =\sigma(\theta_1\mathbf{x}^m + \theta_2)$ with $\boldsymbol{\theta}=(\theta_1, \theta_2)$, at layer $l$ we want to interpolate $f (\mathbf{x}^m;\boldsymbol{\theta})$ such that
\begin{equation}
    f^l(\mathbf{x}^m; \boldsymbol{\theta})=\sigma^l(\theta_1^lD^l f^{l-1}(\mathbf{x}^m) + \theta^l_2)
    \label{eq:emriskg3}
\end{equation}
 predicts the label $y^m$ for $l=2,3,4,\cdots$. In this notation, $f^l$ is the output interpolation of layer $l$, $\sigma^l$ is the activation function at layer $l$, $\boldsymbol{\theta}^l=(\theta^l_1, \theta^l_2)$ is the learnable parameters at layer $l$ with $\theta^l_1$ and $\theta^l_2$ the weight matrix and bias vector respectively. In the expression in Equation~\ref{eq:emriskg3} the weights matrix $D^l$ is introduced as a sort of regularization that activates the connections which  contribute to the interpolation of $f^l(\mathbf{x}^m; \boldsymbol{\theta})$ at layer $l$; this is known as the dropout regularization.

Adopting a gradient flow training method with variable learning rate $\alpha_l$ at layer $l$, in the meta-learning regime as we adopted in this work, the update of $\boldsymbol{\theta}$  follows two procedures. If $p(\mathcal{T})$ is assumed to be the distribution of tasks where each task is sampled as $\mathcal{T}_i\sim p(\mathcal{T})$ with the aim to learn prior knowledge from all these $\mathcal{T}_i$. As discussed in~\cite{finn2017model} the main goal is to encapsulate the prior knowledge of all $\mathcal{T}_i$ as the initial weight $\boldsymbol{\theta}$ of the fitted function $f(\mathbf{x},\boldsymbol{\theta})$ which can now be used as an initial weight for quick adaptation to a new task.
The first attempts is to find the parameter $\boldsymbol{\theta}_{i,k}$ of a task $\mathcal{T}_i$ with training sample $\mathcal{D}_i=\{(\mathbf{x}^{m},y^{m})^i\}$; $m=1,\cdots, N_i$ where $N_i$ is the number of sample in $\mathcal{D}_i$. At the $(k+1)^{\mathrm{th}}$ iteration, $\boldsymbol{\theta}_{i,k}$ is updated as:
\begin{equation}
   \boldsymbol{\theta}_{i,k+1} = \boldsymbol{\theta}_{i,k}-\alpha^l \nabla_{\boldsymbol{\theta}}\sum_{\mathcal{D}_i}\frac{1}{N_i}\mathcal{L}_{\mathcal{T}_i}(y^m, f(\mathbf{x}^{m}; \boldsymbol{\theta}_{i,k})), ~\boldsymbol{\theta}_{i,0}=\boldsymbol{\theta}
   \label{eqO}
\end{equation}
which is now followed  by a proper update of $\boldsymbol{\theta}$ using  the direction of the gradient and the test samples $\mathcal{D}_i^\prime=\{(\mathbf{x}^{m},y^{m})^i\}$ of the task $\mathcal{T}_i$; $m=1,\cdots, N_i^\prime$ where $N_i^\prime$ is the number of sample in $\mathcal{D}_i^\prime$. Assume that $\boldsymbol{\theta}_{i}^\prime$ is obtained after several update as discussed in Equation~\ref{eqO} for each task $\mathcal{T}_i$, the proper update of $\boldsymbol{\theta}_{}$ follows:
\begin{equation}
   \boldsymbol{\theta}_{}  \gets	\boldsymbol{\theta}-\beta^l \nabla_{\boldsymbol{\theta}}\sum_{\mathcal{T}_i}\sum_{\mathcal{D}_i^\prime}\frac{1}{N_{\mathcal{T}} N_i^\prime}\mathcal{L}_{\mathcal{T}_i}(y^m, f(\mathbf{x}^{m}; \boldsymbol{\theta}_{i}^\prime)),
\end{equation}
where $N_{\mathcal{T}}$  and $\beta^l$ are the number of tasks and the learning rate at layer $l$ respectively. 
\subsection{Proposed Visual Explainable Framework}
\label{proposemethod}
\begin{figure}
  \centering
  \includegraphics[width=1.0\textwidth, height=250pt]{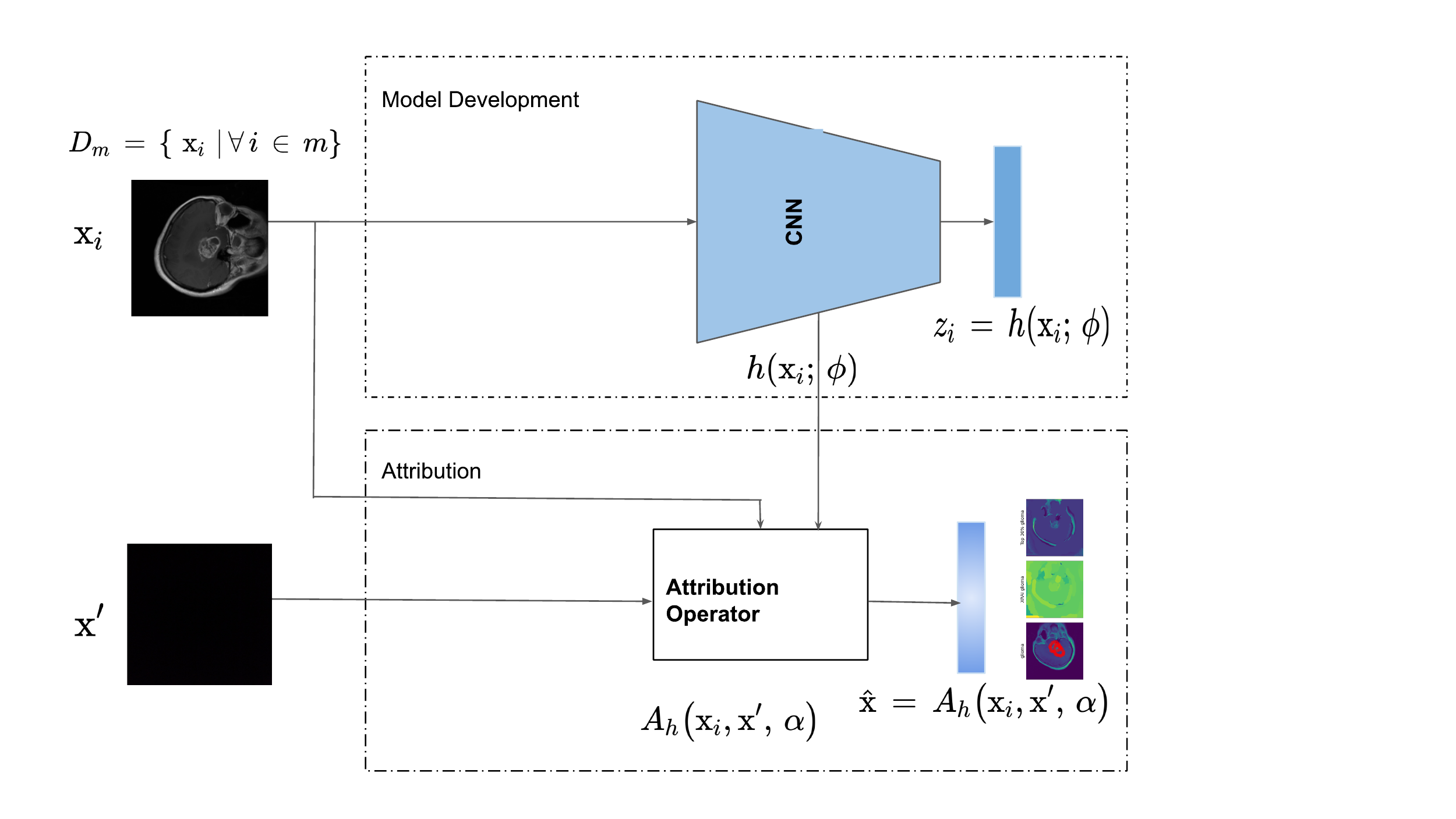}
  \caption{A dataset of $m$  samples of T1-weighted contrast-enhanced images slices is the input to a standard CNN classification model depicted in the figure as $h(\cdot)$ that learns the non-linear mapping of the features to the output labels. $h(\cdot)$ is utilized with an attribution operator $A_h$ to attribute salient features $\hat{\mathbf{x}}$ of the input image. $A_h$ is an operator that can be used with varied differentiable architectures. This proposed framework is general and can be applied to any problem instances where explainability is vital in building trust in the model inference mechanism.}
  \label{fig:system_model_attribution}
\end{figure}

To help interpret a model inference mechanism, which is crucial in building trust for clinical adoption of deep learning-based computer-aided diagnostic systems, we have proposed an interpretability framework depicted in Figure~\ref{fig:system_model_attribution} that gives an overview of an attribution mechanism.~\cite{sundararajan2017axiomatic} posited fundamental axioms: Sensitivity and Implementation Invariance that attribution methods must satisfy. All selected saliency methods in this study adhere to this axiom. For a macro-scale attribution, a model $h(\mathbf{x}_i; \phi)$ that has learned statistical regularities of any given bioimaging dataset $D_{m}$ that has an arbitrary number of classes to produce a representation $z_i$ for each medical image slice $\mathbf{x}_i$ that is a compact latent representation in a vector space. With this representation, any arbitrary dimensionality reduction method can map the latent representation onto a lower-dimensional space for analysis and visualization. This could be a Gaussian Mixture Model (GMM)~\cite{duda1973pattern}, t-Distributed Stochastic Neighbor Embedding (t-SNE)~\cite{van2008visualizing} or Principal Component Analysis (PCA)~\cite{wold1987principal} technique to understand the latent space projection.

To attain local information about an attribution scheme because of the limitations of global attribution as it does not give contextual information of feature importance in the input space. We, therefore, propose the use of gradient information since neural models are differentiable or at least partially differentiable functions. We propose a framework of an adaptive path-based gradient integration method that utilizes the Guided Integrated Gradient (GIG)~\cite{kapishnikov2021guided} as shown in Equation~\ref{eq:ig} and a Region-based saliency method: eXplanation with Ranked Area Integrals (XRAI)~\cite{kapishnikov2019xrai}. 
The core idea of Integrated Gradient (IG) is that given a non-linear differentiable function $h$ defined as:
\begin{align}
     h: &\mathbb{R}^{n} \longrightarrow[0,1]\\
        &\mathbf{x} \longmapsto h(\mathbf{x}),
\end{align}
which represents a deep neural network and an input  $\mathbf{x}=\left(x_{1}, \ldots, x_{n}\right) \in \mathbb{R}^{n}$. A general attribution of the prediction  at the input $\mathbf{x}$ relative to some baseline input $\mathbf{x}^{\prime}$ is a vector $A_{h}\left(\mathbf{x}, \mathbf{x}^{\prime}\right)=\left(a_{1}, \ldots, a_{n}\right) \in \mathbb{R}^{n}$ where $a_{i}$ is the contribution of the vector component $x_{i}$ to the function $h(\mathbf{x})$. In a medical image analysis context, the function $h$ represents a deep neural network that learns a disentangled non-linear transformation of given medical image slices. The input vector $\mathbf{x}$ is a simple tensor of the $k$ channel image, where the indices correspond to pixels. The attribution vector $\mathbf{a}=\left(a_{1}, \ldots, a_{n}\right)$ serves as a mask over the original input to show the regions of interest of the model for the given predicted score. This information gives us insight into regions of interest for any given 2D image slice:
\begin{equation}
I G_{i}(\mathbf{x})=\left(\mathbf{x}_{i}-\mathbf{x}_{i}^{\prime}\right) \int_{\alpha=0}^{\alpha=1} \frac{\partial }{\partial \mathbf{x}_{i}} h\big(\mathbf{x}^{\prime}+\alpha\left(\mathbf{x}-\mathbf{x}^{\prime}\right)\big) \mathrm{d} \alpha,
\label{eq:ig}
\end{equation}
where $\left(\mathbf{x}_{i}-\mathbf{x}_{i}^{\prime}\right)$ is the difference between the input image and the corresponding baseline input at each pixel.

Computing and visualizing the saliency maps involve the following steps: 
\begin{enumerate}
    \item  We initialize a baseline with all zeros. This baseline input remains prediction-neutral and has a crucial role in the interpretation and visualization of the input pixel feature importance.
    \item  Linear interpolations are generated between the baseline and the original image that are incremental steps $(\alpha)$ in the feature space between the baseline $\mathbf{x}^{\prime}$ and the input image $\mathbf{x}$.
    \item The gradient in Equation~\ref{eq:ig} is computed to measure the relation between the features $\mathbf{x}_i$ and changes in the model class predictions. It gives a criterion for pixels with the most relevance to the model class probability scores. This gives a basis for quantifying feature importance in the input image with respect to the model prediction.
    \item Using a summation method, an aggregate of the gradients is computed.
    \item The aggregated saliency mask is scaled to the input image to ensure that feature attribution values are accumulated across multiple interpolated images that are all on the same scale that represents the saliency map on the input image with the pixel feature saliency.
\end{enumerate}
\section{Experimental Results}
\label{results}
In this section, we present an overview of the datasets used in this paper including the annotation procedure for the segmentation of regions of interest in each MRI image. We further explain the training regime for all the models and elaborate on the framework for computing interpretable features using adaptive path-based gradient integration techniques for scoring pixel-wise feature relevance as discussed in Section~\ref{proposemethod}. Results show that deep neural network models trained on medical images can give prediction confidence through softmax scores as well as use visual interpretability techniques to infer feature attribution maps.

\subsection{Datasets}
\begin{figure}
  \centering
  \includegraphics[width=1.0\textwidth, height=340pt]{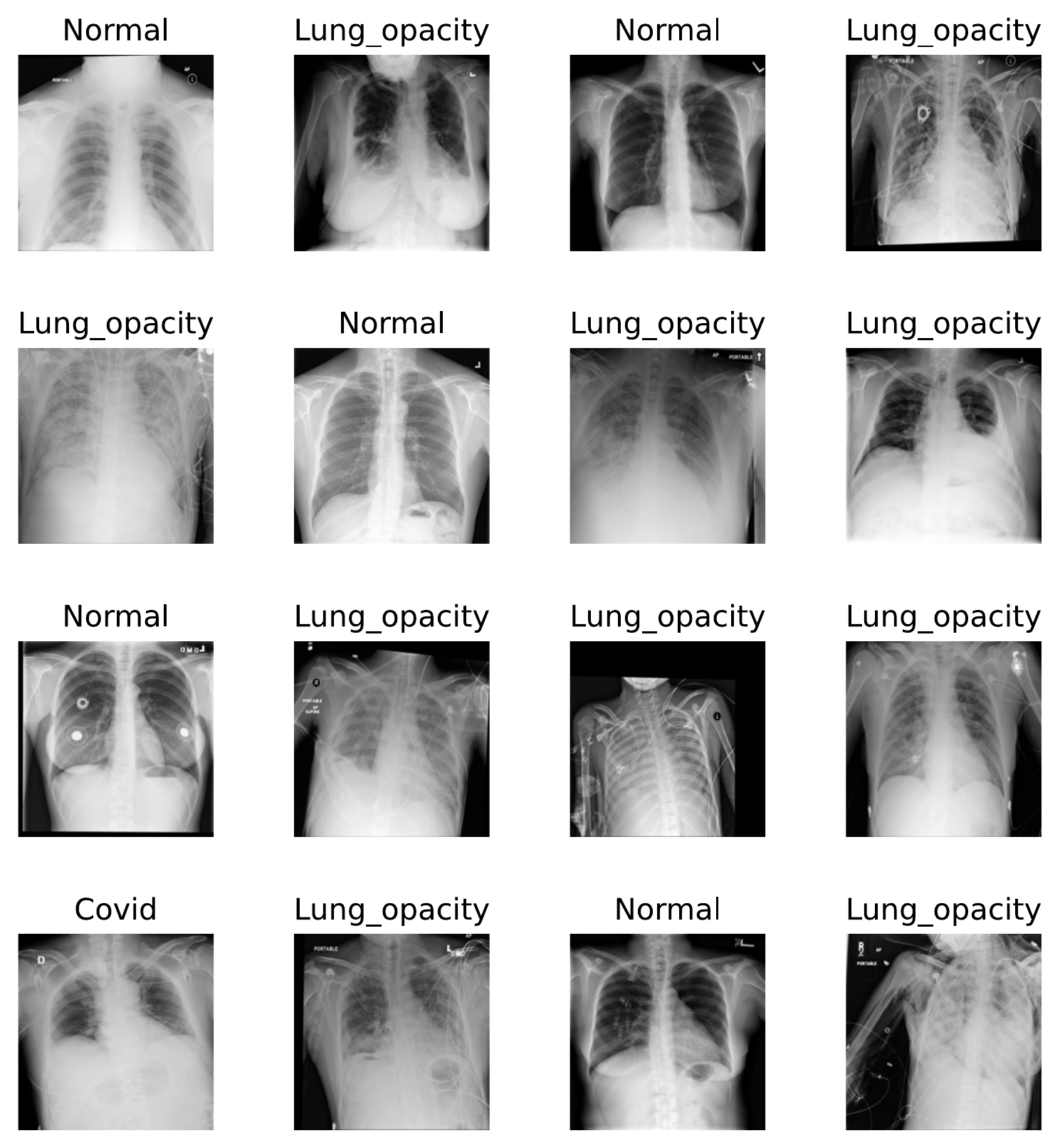}
  \caption{Random selected 16 samples. The dataset was curated from multiple international COVID-19 X-ray testing centers during several time periods. The dataset is made up of an unbalanced percentage of the four classes in which we have 48.2 $\%$ normal X-ray images, 28.4 $\%$ cases with lung opacity, 17.1 $ \%$ of COVID-19 patients and $6.4\%$ of patients with viral pneumonia of the 19820  total images in the dataset. The highly unbalanced percentages explain the occurrence of normal and lung opacity cases in the random selection versus COVID-19 and/or viral pneumonia.}
  \label{fig:COVID-19_Radiography_Dataset}
\end{figure}

We use two types of medical image data modalities to test the attribution framework. The choice of the two modalities depends on the availability of data. Other types of modalities are also applicable to the attribution framework. We leave this for future work.
\begin{figure}
  \centering
  \includegraphics[width=1.0\textwidth, height=335pt]{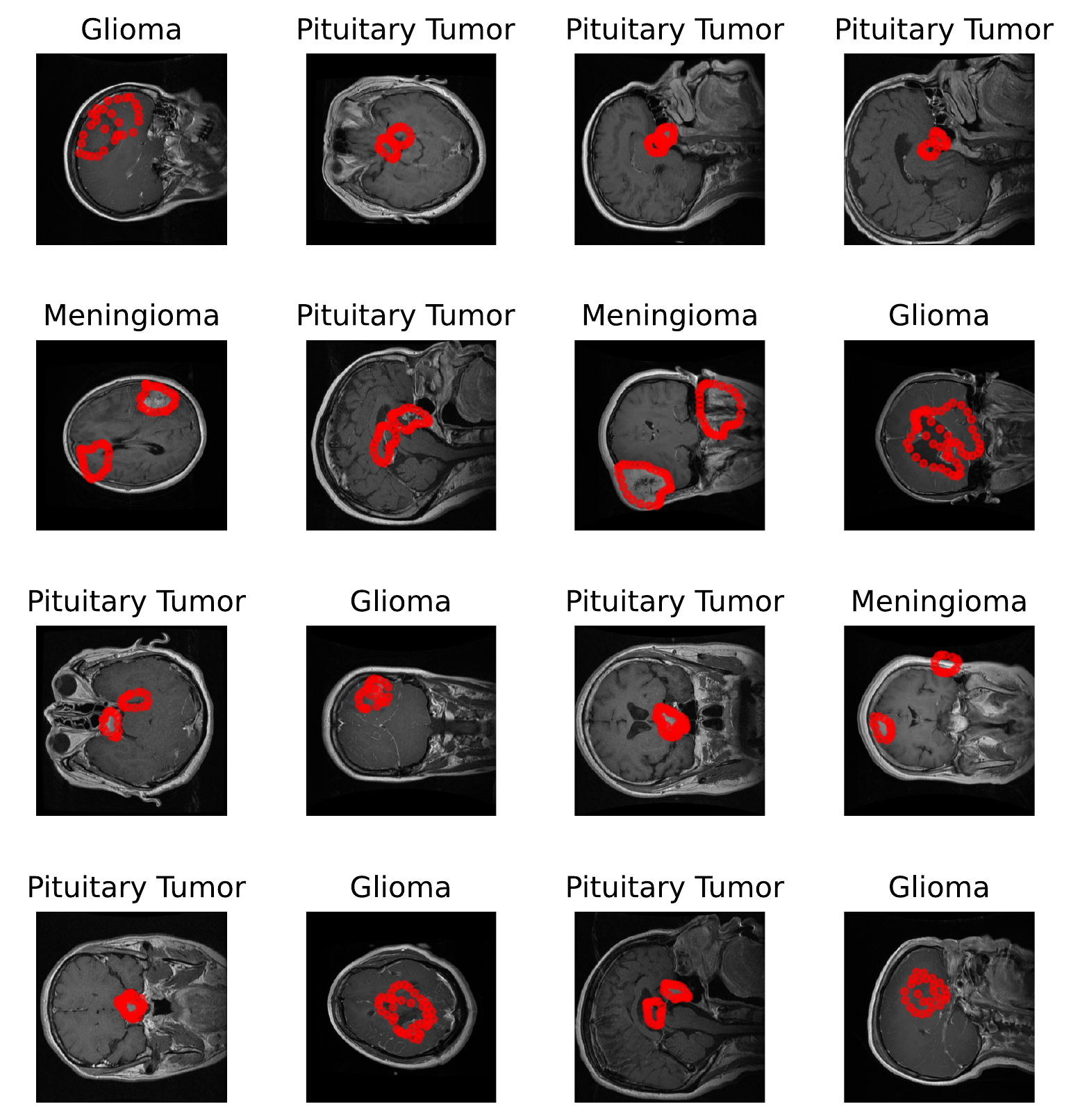}
  \caption{ Shows randomly sampled images from the brain tumor dataset. The red annotated regions indicate perimeters of segmented tumor borders. From the figure, Glioma samples have the widest tumor areas as opposed to the other two tumor classes. Glioma tumor tissue can be formed in varied locations in the brain. Like Glioma, a Meningioma is a primary central nervous system (CNS) tumor and can begin in the brain or spinal cord areas. Meningioma is the most common type of tumor among patients. As shown in the figure, samples often occur in pairs across opposite regions of the brain. As depicted in the figure, Pituitary tumors are abnormal growths that develop in the pituitary gland that lead to excess hormonal releases that regulate important body functions.}
  \label{fig:Samples}
\end{figure}
The brain tumors MRI dataset~\cite{Cheng2017} is used. It comprises 2D slices of brain contrast-enhanced MRI (CE-MRI) T1-weighted images consisting of 3064 slices from 233 patients. It includes 708 Meningiomas, 1426 Gliomas, and 930 Pituitary tumors. Representative MRI image slices with large lesion sizes are selected to construct the dataset. In each slice, the tumor boundary is manually delineated and verified by radiologists. We have plotted 16 random samples from the three classes with tumor borders depicted in red as shown in Figure~\ref{fig:Samples}. These 2D slices of T1-weighted images train standard deep CNNs for a 3-class classification task into Glioma, Meningioma, and Pituitary tumors. The input to each model is a $\mathbb{R}^{225\times225\times1}$ tensor that is a resized version of the original $\mathbb{R}^{512\times 512}$ image slices primarily due to computational concerns.
\begin{table}[!ht]
\caption{The 2 datasets comprising different modalities used to carry out experiments in this study.}
\label{tab:datasets}
\centering
\begin{tabular}{l l l l l l}
\toprule
\multicolumn{1}{c}{\textbf{Source}} & \multicolumn{1}{c}{\textbf{Classes}} & \multicolumn{1}{c}{\textbf{\begin{tabular}[c]{@{}c@{}}Number \\ of samples\end{tabular}}} & \multicolumn{1}{c}{\textbf{Total}} & \multicolumn{1}{c}{\textbf{Modality}} & \multicolumn{1}{c}{\textbf{Segmented}} \\ \cmidrule(lr){1-6}
\multirow{3}{*}{Brain Tumor Dataset~\cite{Cheng2017}} & Meningioma & 708 & \multirow{3}{*}{3064} & \multirow{3}{*}{MRI} & \multirow{3}{*}{yes} \\ 
 & Glioma & 1,426 &  &  &  \\ 
 & Pituitary tumor & 930 &  &  &  \\ 
\multirow{3}{*}{COVID-19 database~\cite{9144185}} & COVID-19 & 3,616 & \multirow{3}{*}{19,820} & \multirow{3}{*}{X-ray} & \multirow{3}{*}{no} \\ 
 & Normal & 10,192 &  &  &  \\ 
 & Lung Opacity & 6,012 &  &  &  \\ 
 \bottomrule
\end{tabular}
\end{table}
Unlike the brain cancer MRI dataset which comes with segmentation masks from experts in the field, the COVID-19 X-ray dataset~\cite{chowdhury2020can} used in this work has no ground truth segmentation masks. This was chosen as an edge-case analysis due to the fact that a vast majority of datasets do not have segmentation masks. This dataset was curated from multiple international COVID-19 X-ray testing facilities during several time periods. The dataset is made up of an unbalanced percentage of the four classes in which we have 48.2 $\%$ normal X-ray images, 28.4 $\%$ cases with lung opacity, 17.1 $ \%$ of COVID-19 patients and $6.4\%$ of patients with viral pneumonia of the 19820  total images in the dataset. This unbalanced nature of the dataset comes with its own classification challenges and has prompted several researchers to implement methods to classify the dataset using deep learning methods. Out of the four classes, for consistency with the other datasets used in this work, we choose to classify three classes (i.e., Normal, Lung Opacity, and COVID-19). For an in-depth discussion of works that deal with this dataset, we refer to~\cite{diagnostics11081480}. Figure~\ref{fig:COVID-19_Radiography_Dataset} shows 16 selected random samples. Table~\ref{tab:datasets} summarizes those three datasets.

\subsection{Implementation Performance}
\begin{figure}
     \centering
     \begin{subfigure}[b]{0.48\textwidth}
         \centering
           \includegraphics[width=1.0\textwidth, height=150pt]{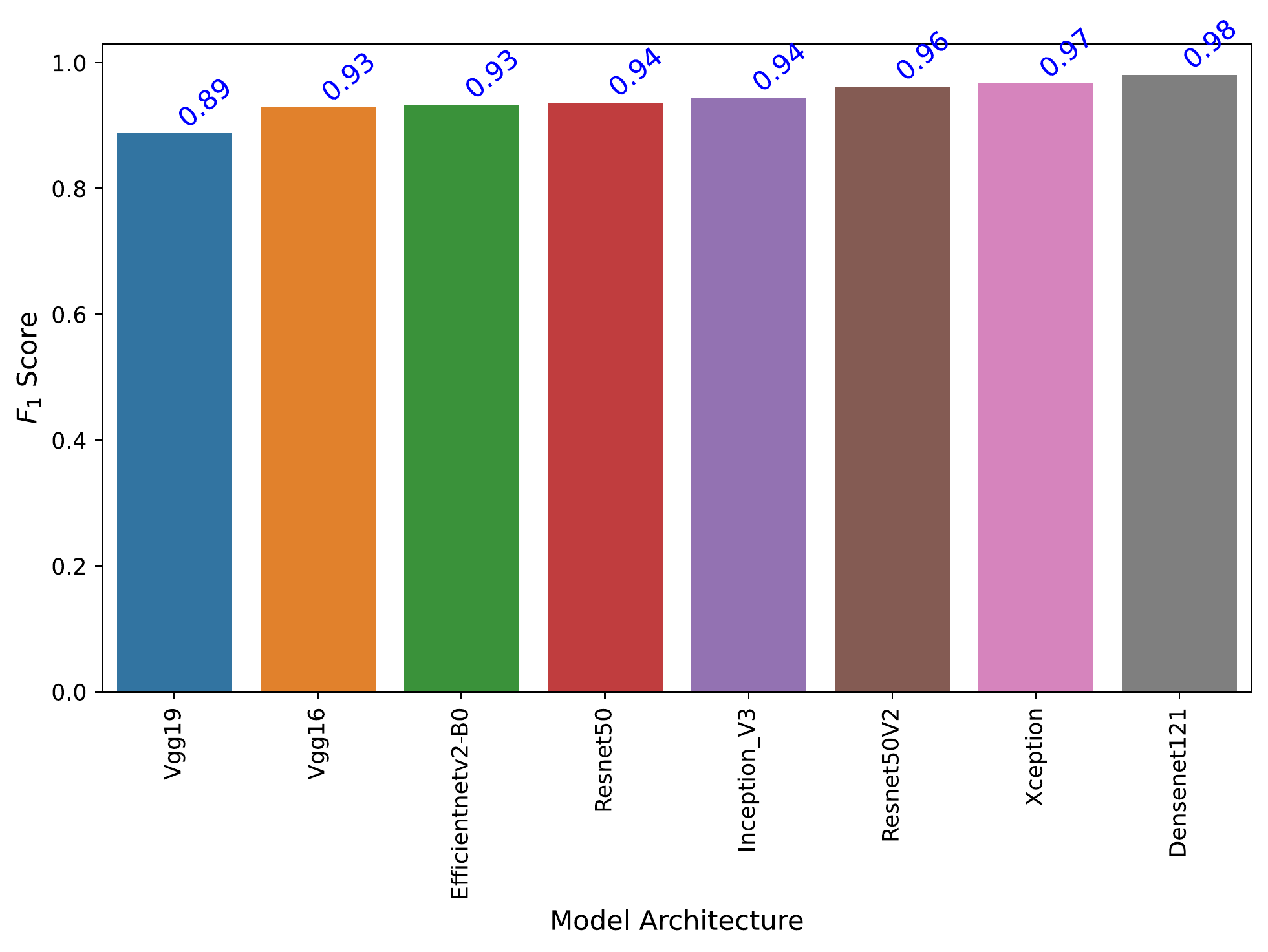}
         \label{fig:Train_Time_Accuracy}
     \end{subfigure}
     \hfill
     \begin{subfigure}[b]{0.48\textwidth}
         \centering
           \includegraphics[width=1.0\textwidth, height=150pt]{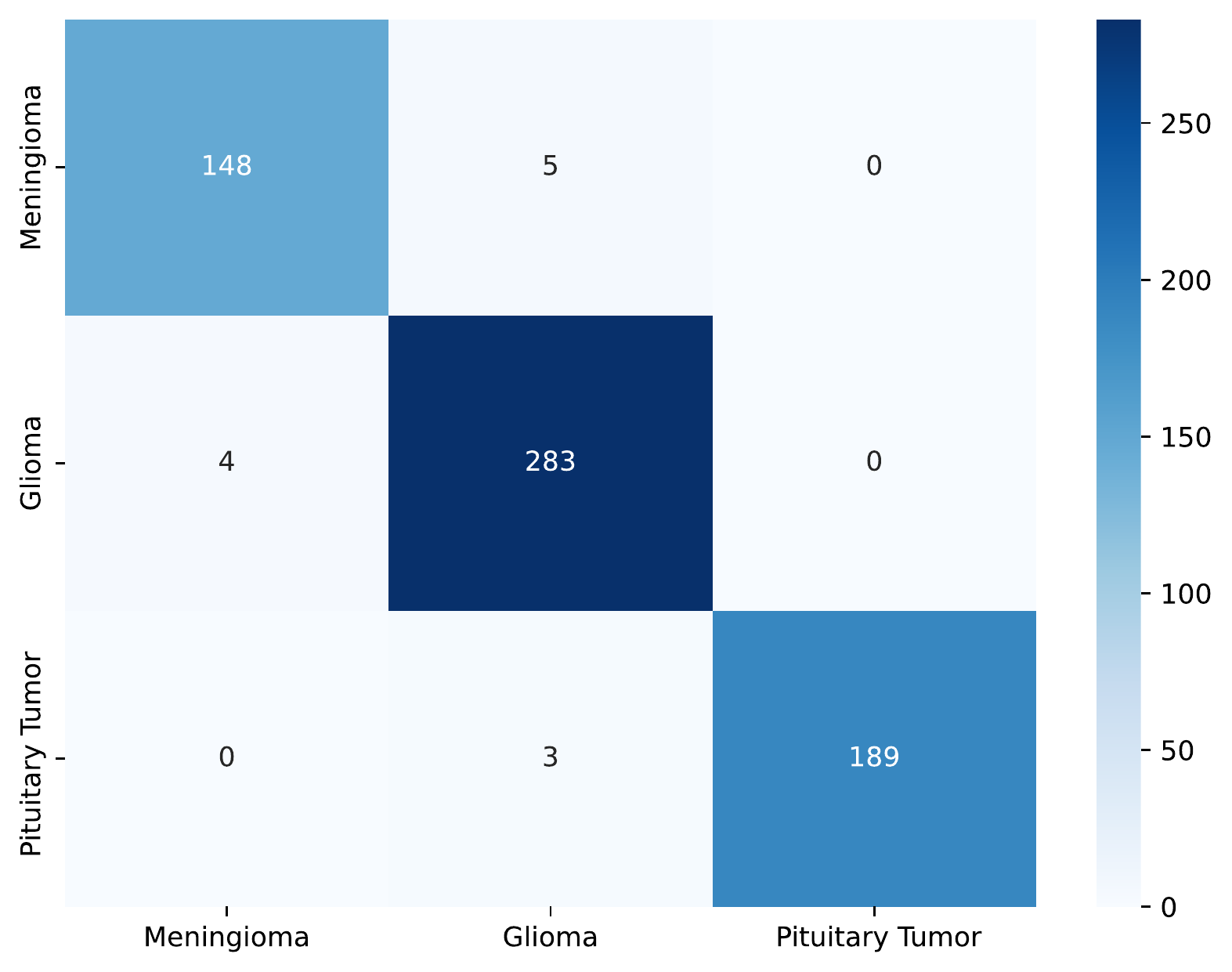}
         \label{fig:densenet121_plot_confusion_matrix}
     \end{subfigure}
        \caption{Performance measure of the 8 CNN architectures used in this experiment all trained for 20 epochs on the brain MRI dataset. Overall, DenseNet121~\cite{chollet2017xception} showed the highest $F_1$ Score reaching 0.981. The confusion matrix for test samples represents 10\% of the dataset. The model could generalize well with 5, 4, and 3 misclassifications for Meningioma, Glioma, and Pituitary tumor respectively. Because of the distinctness of both Meningioma and Pituitary tumor, the model has 0 false positives between both classes.}
        \label{fig:densenet121_plot_confusion_matrix_combined}
\end{figure}

\begin{figure}
     \centering
     \begin{subfigure}[b]{0.48\textwidth}
         \centering
           \includegraphics[width=1.0\textwidth, height=150pt]{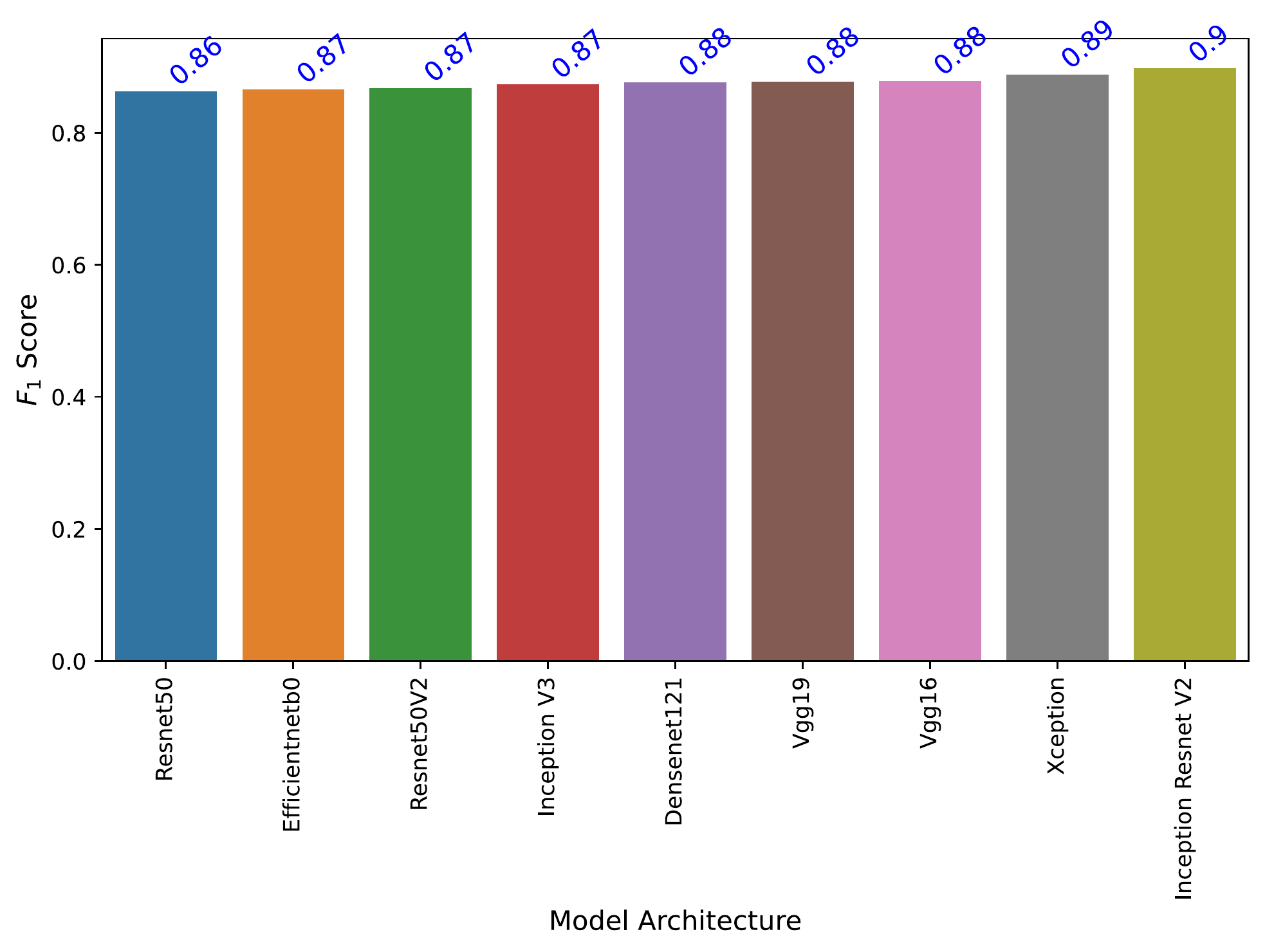}
         \label{fig:19_Radiography_Dataset_Train_Time_Accuracy}
     \end{subfigure}
     \hfill
     \begin{subfigure}[b]{0.48\textwidth}
         \centering
           \includegraphics[width=1.0\textwidth, height=150pt]{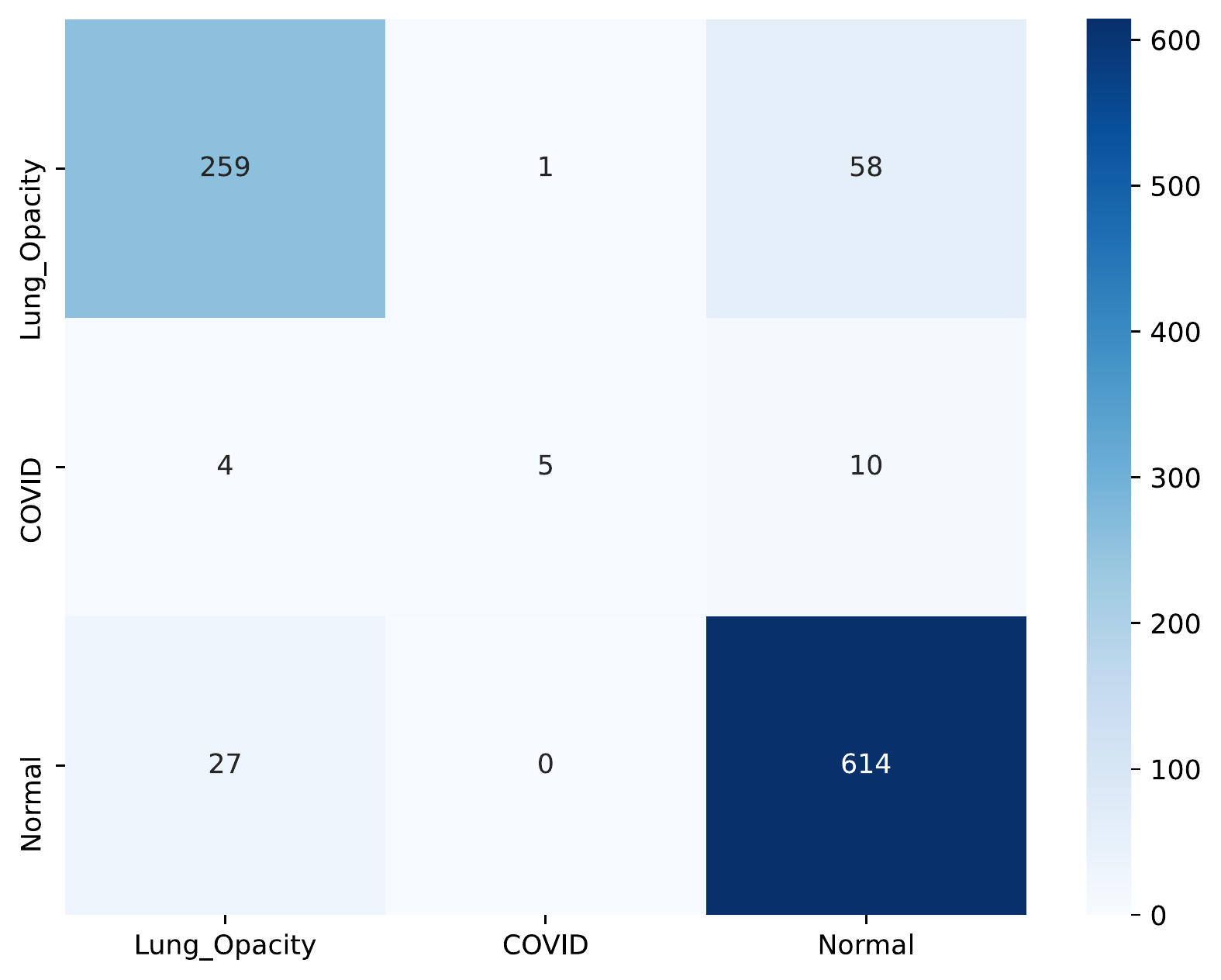}
         \label{fig:19_Radiography_Dataset_inception_resnet_v2_plot_confusion_matrix}
     \end{subfigure}

        \caption{InceptionResNetV2 reached the best test-time performance for the chest X-ray dataset. All models nearly uniformly performed well on this dataset primarily because of the huge number of data points that are well-suited for high-capacity models to prevent overfitting. From the corresponding confusion matrix, on the left, Lung Opacity has the largest number of misclassification relative to the distribution of the dataset.}
        \label{fig:19_Radiography_Dataset_inception_resnet_v2_plot_confusion_matrix_combined}
\end{figure}

 As the primary objective of this study is to build a framework for understanding the visual interpretability of deep learning models in medical image analysis, we limit our experiments to 9 modern vision-based deep neural architectures.  We trained and tested the 9 modern CNN architectures; results are shown in Figures~\ref{fig:densenet121_plot_confusion_matrix_combined}, and~\ref{fig:19_Radiography_Dataset_inception_resnet_v2_plot_confusion_matrix_combined} and summarized in Table~\ref{tab:Model_Architectures} with training hyperparameters depicted in Table~\ref{Tab:ResultSummary} for the two datasets used to test the proposed attribution method. The object of this work is not to find models that outperform the current literature with the different datasets, but rather to answer the question: what do the deep learning models learn in medical images via the proposed attribution  method? 
 We conducted all experiments on an NVIDIA K80/T4 GPU.  In Section~\ref{attributionxx} several saliency methods are applied to understand model prediction interpretability.
\begin{table}[!htp]
\centering
\caption{A comparison of the 9 models on the test set including their architectural properties. DenseNet121 has the best overall performance on the unseen test set reaching a top-1 accuracy of 98.10\% on the brain tumor MRI dataset. Relative to the least performing model, VGG19, it is not only parameter efficient but has a small memory footprint of at least 16 times less than VGG19. From this table, we chose the top three best-performing models per dataset for saliency analysis considering the impact of parameter count and depth on the type of representations learnable from these models. InceptionResNetV2 outperformed other models for the COVID-19  chest X-ray dataset.}
\label{tab:Model_Architectures}
\begin{tabular}{l l l l l l}
\toprule
\multicolumn{1}{c}{\multirow{2}{*}{\textbf{Model}}} &
  \multicolumn{1}{c}{\multirow{2}{*}{\textbf{Size (MB)}}} &
  \multicolumn{1}{c}{\multirow{2}{*}{\textbf{Parameters}}} &
  \multicolumn{1}{c}{\multirow{2}{*}{\textbf{Depth}}} &
  \multicolumn{2}{c}{\textbf{Top-1 Accuracy}} \\  \addlinespace[5pt] \cmidrule(lr){1-6} 
\multicolumn{1}{c}{} &
  \multicolumn{1}{c}{} &
  \multicolumn{1}{c}{} &
  \multicolumn{1}{c}{} &
  \multicolumn{1}{c}{\begin{tabular}[c]{@{}c@{}}Brain Tumor\\  Dataset\end{tabular}} &
  \multicolumn{1}{l}{\begin{tabular}[c]{@{}l@{}}COVID-19 \\ database\end{tabular}} \\ \cmidrule(lr){5-6} 
VGG16             & 528         & 138.4M        & \textbf{16} & \multicolumn{1}{l}{0.928797}          & \multicolumn{1}{l}{\textbf{0.891616}}  \\
VGG19             & 549         & 143.7M        & 19          & \multicolumn{1}{l}{0.887658}          & \multicolumn{1}{l}{0.889571}           \\ 
ResNet50          & 98          & 25.6M         & 107         & \multicolumn{1}{l}{0.936709}          & \multicolumn{1}{l}{0.857873}           \\ 
ResNet50V2        & 98          & 25.6M         & 103         & \multicolumn{1}{l}{0.962025}          & \multicolumn{1}{l}{0.881391}           \\ 
InceptionV3       & 92          & 23.9M         & 189         & \multicolumn{1}{l}{0.944620}          & \multicolumn{1}{l}{0.880368}           \\ 
Xception          & 88          & 22.9M         & 81          & \multicolumn{1}{l}{0.966772}          & \multicolumn{1}{l}{0.889571}           \\ 
EfficientNetB0    & \textbf{29} & 5.3M          & 132         & \multicolumn{1}{l}{0.933544}          & \multicolumn{1}{l}{0.880368}           \\ 
DenseNet121       & 33          & \textbf{8.1M} & 242         & \multicolumn{1}{l}{\textbf{0.981013}} & \multicolumn{1}{l}{0.884458}           \\ 
InceptionResNetV2 & 215         & 55.9M         & 449         & \multicolumn{1}{l}{-}                 & \multicolumn{1}{l}{\textbf{0.895706}}  \\ \bottomrule
\end{tabular}
\end{table}

\begin{figure}
     \centering
            \includegraphics[width=1.\textwidth]{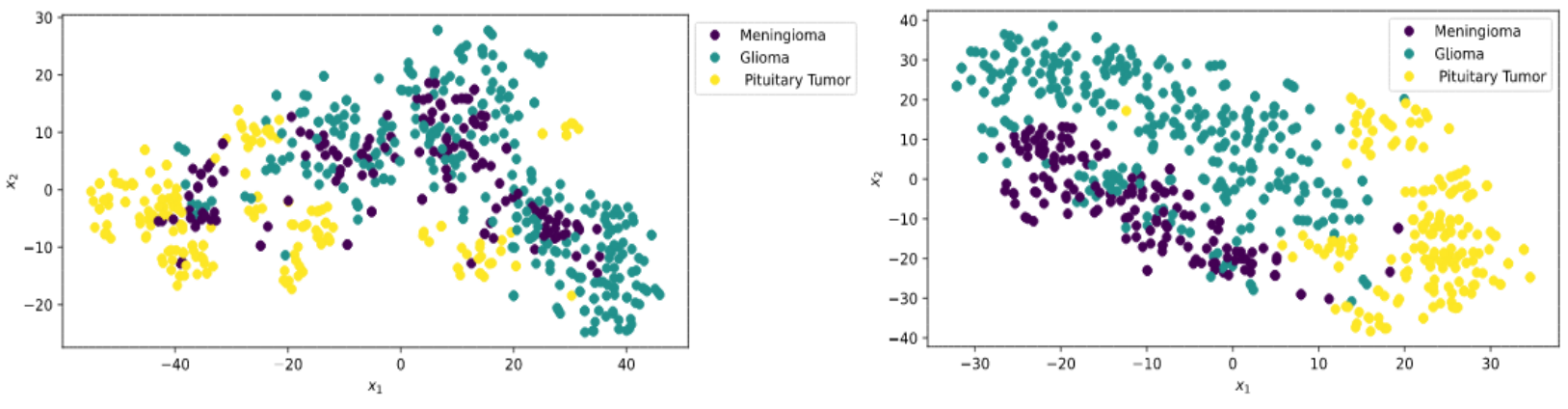}
        \caption{A t-SNE~\cite{van2008visualizing} two-dimensional projection of the unrolled pixel space representation of MRI slices where the colors purple, green, and yellow represent the three classes of  Meningioma, Glioma, and Pituitary tumor respectively. However, given that the data is generated under differing physical and statistical conditions, the classes are entangled. This can impede learning using linear function approximations. (Right) A t-SNE projection of the embedding representation from a trained DenseNet121 network. The model has disentangled the underlying factors of variation in a latent representation space that allows separability using either linear or non-linear function approximators as shown by the nearly distinct manifolds of the three classes. }
        \label{fig:Test_set_xception_combined}
\end{figure}

\begin{figure}
     \centering
      \includegraphics[width=1.\textwidth]{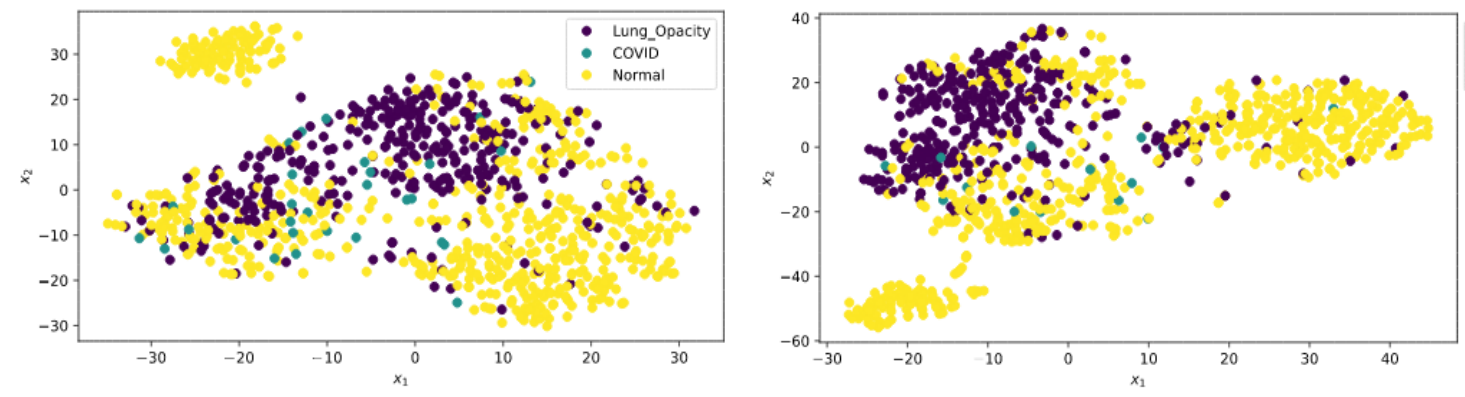}
        \caption{A similar 2D t-SNE visualization of the InceptionResNetV2 latent representations for the chest X-ray dataset. This dataset has a rich statistical structure across all classes, however, it is also imbalanced like many medical datasets. (Right) A plot of latent embeddings prior to training, the dataset is biased towards normal class which was addressed through class weighting during training. (Right) Embeddings of the network after training. There is a visible decrease in the intra-class cluster size as samples belonging to the same class are pulled closer during the training phase in the representation space. This notion is supported by the confusion matrix plot in Figure~\ref{fig:19_Radiography_Dataset_inception_resnet_v2_plot_confusion_matrix_combined}.} 
    \label{fig:COVID-19_Radiography_Dataset_Test_set_COVID-19_Radiography_Dataset_inception_resnet_v2_Combined}
\end{figure}

\begin{table}[!htp]
\centering
\caption{\label{tab:hyperparameters}Training hyperparameters. }
\label{Tab:ResultSummary}
\begin{tabular}{ll}
\toprule Hyperparameter & Setting \\
\cmidrule(lr){1-2}
Learning rate & 1e-3\\
Batch size & 32\\
Number of epochs & 20\\
Training set & 0.7 \\
Test set & 0.3\\
Input shape & $\mathbb{R}^{225 \times 225 \times 1}$\\
Momentum & 9.39e-1\\
Decay  &  3e-4\\
Optimizer & Stochastic Gradient Descent with Momentum (SDGM)\\
 \bottomrule
\end{tabular}
\end{table}
 With the brain MRI dataset, the DenseNet121 model shows the best overall test performance reaching 98.10\%. While the hybrid InceptionResNetV2 outperformed the other models on the COVID-19 X-ray dataset with an accuracy of $89.0\%$. The test results indicate the high confidence and stability of model prediction. This is the basis of selection for further feature attribution given that it is the best-performing model implying it has learned a more robust and generalizable representation of the data distribution as shown in Figures~\ref{fig:Test_set_xception_combined}, and~\ref{fig:COVID-19_Radiography_Dataset_Test_set_COVID-19_Radiography_Dataset_inception_resnet_v2_Combined}. The clear distinction between Figures~\ref{fig:Test_set_xception_combined}, and~\ref{fig:COVID-19_Radiography_Dataset_Test_set_COVID-19_Radiography_Dataset_inception_resnet_v2_Combined} left and  right panels give an evident indication that the model has learned inherent factors of variation in the signals which have been disentangled into nearly separable manifolds in the learned representation space). These figures support the results of the confusion matrices in Figures~\ref{fig:densenet121_plot_confusion_matrix_combined}, and ~\ref{fig:19_Radiography_Dataset_inception_resnet_v2_plot_confusion_matrix_combined}. However, this ability of learning necessitates the notion of what has the model learned about the data space and how can it be interpreted by domain experts. Thus, the notion of feature attribution is investigated to make sense of mapping between the model input and the predicted class.

\subsection{Attribution}
\label{attributionxx}
\begin{figure}
 \vspace{-1.7cm}
\begin{subfigure}{.5\textwidth}
 \hspace{+3.3cm}
\centerline{\includegraphics[width=.890\paperwidth, height=165pt]{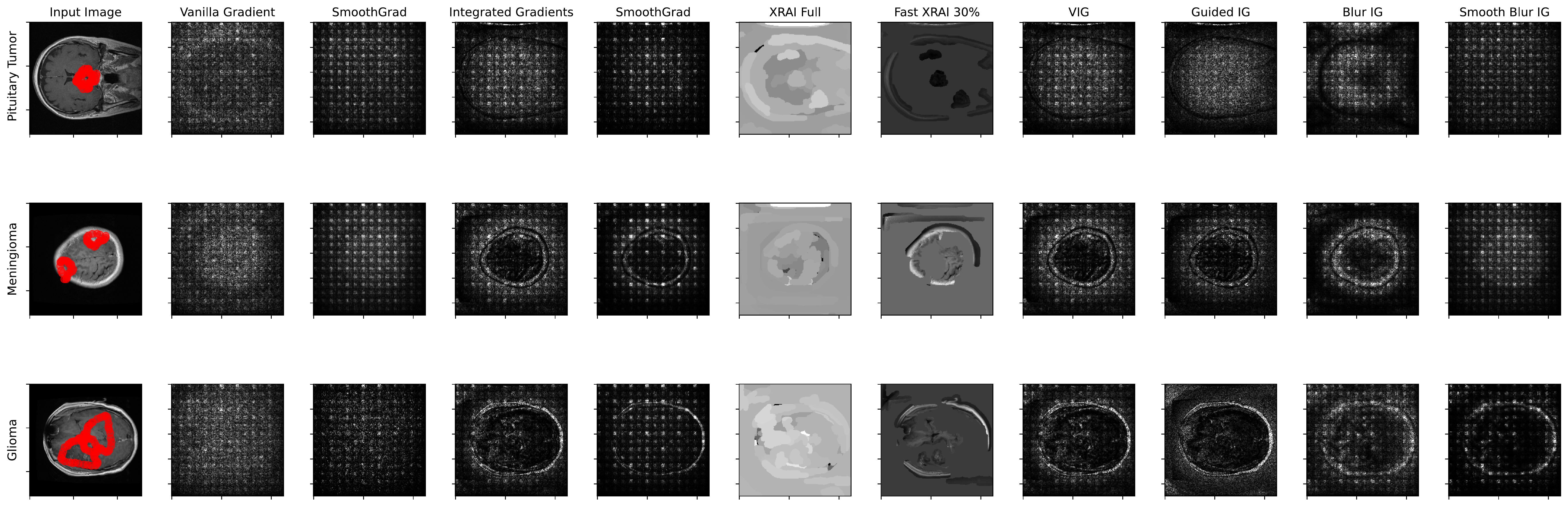}}
   \caption{Xception}
     \label{fig:Saliency_Maps_xception}
\end{subfigure} \\
\begin{subfigure}{.5\textwidth}
 \hspace{+3.3cm}
 \centerline{\includegraphics[width=.890\paperwidth, height=165pt]{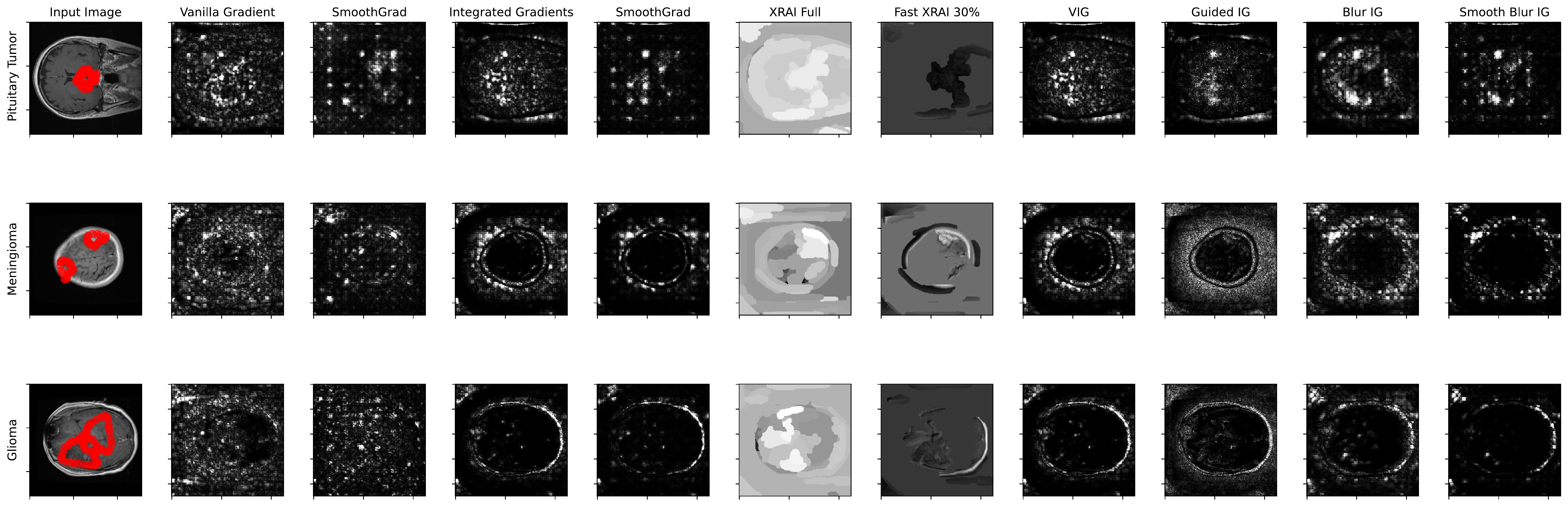}}
  \caption{ResNet50V2}
     \label{fig:Saliency_Maps_resnet50v2}
\end{subfigure} \\
\begin{subfigure}{.5\textwidth}
 \hspace{+3.3cm}
\centerline{\includegraphics[width=.890\paperwidth, height=165pt]{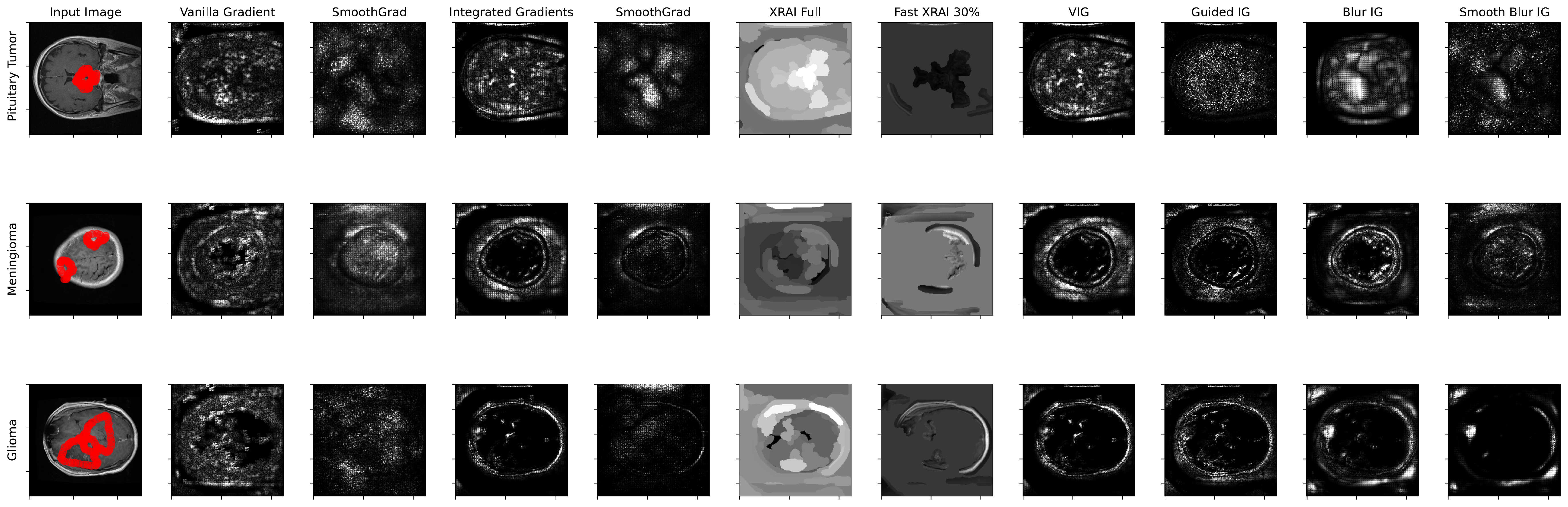}}
    \caption{DenseNet121}
      \label{fig:Saliency_Maps_densenet121}
\end{subfigure} 
 \caption{Brain tumor MRI: In the first column on the left is the input image where the red borders depict the delineated boundaries of tumors.  Three randomly sampled test images from each tumor class are chosen for saliency analysis using the top 3 trained models.}
   \label{MRI:attribution}
\end{figure}

\begin{figure}
 \vspace{-1.8cm}
\begin{subfigure}{.5\textwidth}
 \hspace{+3.3cm}
\centerline{\includegraphics[width=.89\paperwidth, height=165pt]{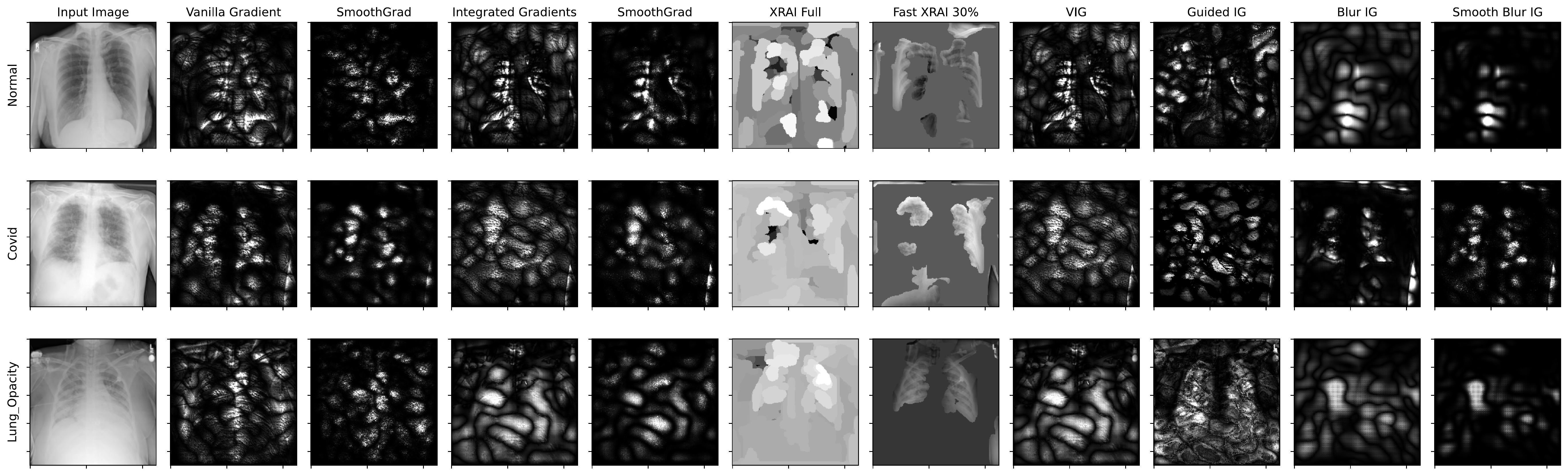}}
  \caption{VGG16}
  \label{fig:COVID-19_Radiography_Dataset_Saliency_Maps_vgg16}
\end{subfigure} \\
\begin{subfigure}{.5\textwidth}
 \hspace{+3.3cm}
 \centerline{\includegraphics[width=.890\paperwidth, height=165pt]{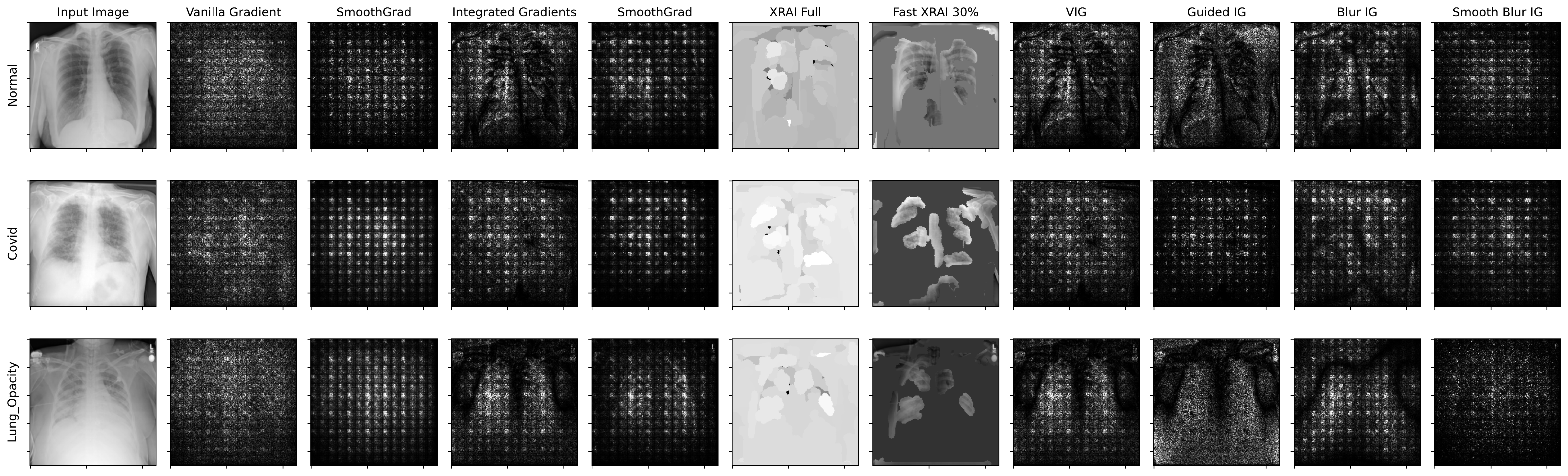}}
  \caption{Xception}
  \label{fig:COVID-19_Radiography_Dataset_Saliency_Maps_xception}
\end{subfigure} \\
\begin{subfigure}{.5\textwidth}
 \hspace{+3.3cm}
\centerline{\includegraphics[width=.89\paperwidth, height=165pt]{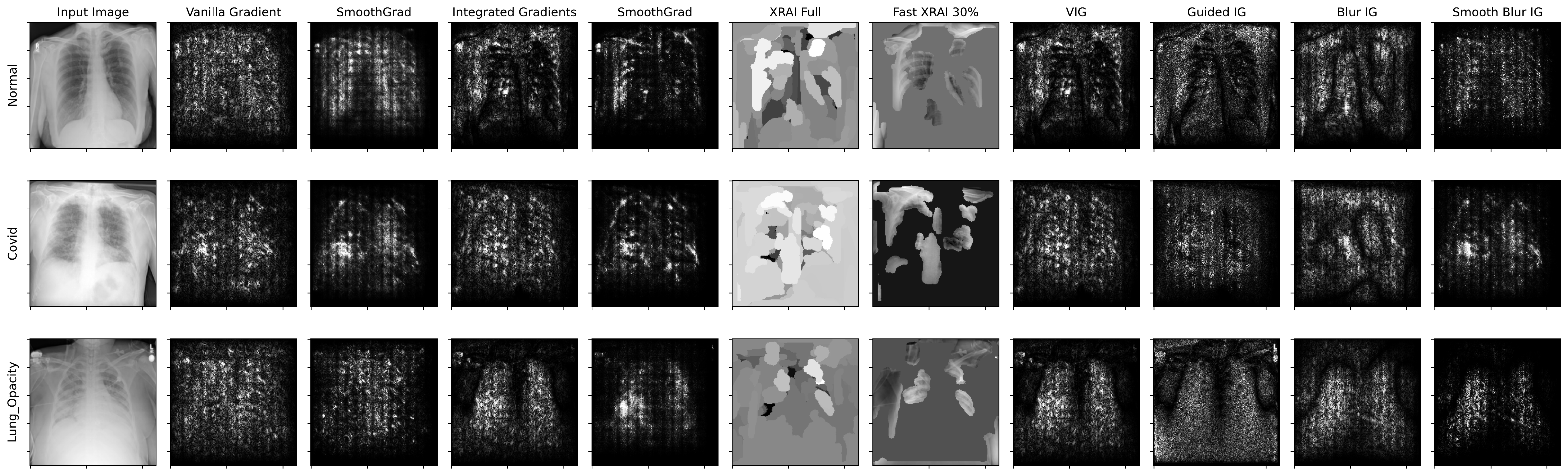}}
  \caption{InceptionResNetV2}
  \label{fig:COVID-19_Radiography_Dataset_Saliency_Maps_inception_resnet_v2}
\end{subfigure} 
 \caption{COVID-19 X-ray:  Three randomly sampled test images from each tumor class are chosen for saliency analysis using the trained (a) VGG16, (b) Xception, and (c) InceptionResNetV2. The infected regions are not segmented from the studied dataset.}
   \label{COVID:attribution}
\end{figure}

Our proposed framework for understanding attribution is predicated on the notion that visual inspection has a major role in medical image analysis decision-making. Naturally, an automated visual attribution method is vital in a human-centered AI medical image analysis pipeline. Given that many attribution methods have been proposed, we have, however, used gradient-based adaptive path integration methods because of their robustness to noise and smoother pixel-level feature saliency mappings. For each of  the datasets, the proposed  visual attribution framework is implemented with the Vanilla Gradient~\cite{sundararajan2017axiomatic}, Guided Integrated Gradient (GIG)~\cite{kapishnikov2021guided} and XRAI~\cite{kapishnikov2019xrai} using the three best performing deep learning models for each dataset as shown in Figures~\ref{fig:densenet121_plot_confusion_matrix_combined}, and~\ref{fig:19_Radiography_Dataset_inception_resnet_v2_plot_confusion_matrix_combined}; i.e. DenseNet121, Xception, and ResNet50V2 are the best three models for  the brain tumor MRI dataset and Inception-ResNetV2, Xception, and VGG16 for the COVID-19 X-ray dataset. Results are depicted in Figures~\ref{fig:Saliency_Maps_xception},~\ref{fig:Saliency_Maps_resnet50v2},~\ref{fig:Saliency_Maps_densenet121}  for the three brain MRI tumor classes and in Figures~\ref{fig:COVID-19_Radiography_Dataset_Saliency_Maps_vgg16},~\ref{fig:COVID-19_Radiography_Dataset_Saliency_Maps_xception}, and ~\ref{fig:COVID-19_Radiography_Dataset_Saliency_Maps_inception_resnet_v2} for three COVID-19 X-ray classes.

Figures~\ref{MRI:attribution}, and~\ref{COVID:attribution} shows three randomly sampled test images from each brain tumor MRI, and chest X-ray that are chosen for saliency analysis using the three trained best deep learning models for each of the datasets.  Each of the image modalities undergoes saliency analysis using each of the attribution methods as shown in the first row titles from Vanilla Gradient-based to Smooth Blur Guided Integrated Gradients. The images are plotted on a grayscale, and the bright spots for the brain tumor show the regions in the input selected for classification into the predicted class by the model. Overall, XRAI has the best explainability of the input signals. This is further explored by pruning 30\% of less explainable features of the attributed image as presented in the Fast XRAI 30\%. There is an emergence of salient features that correspond to the input region of interest for each tumor class. 
In contrast to other deep learning models,  the saliency maps of the Xception model have the least saliency map stability with increased noise levels across all three brain MRI classes. More importantly, XRAI has wider regions of interest computed that correspond to the input signal segmentation mask. DenseNet121 and InceptionResNetV2 are the overall best-performing models in this study for the brain tumors, and chest X-ray datasets respectively. This is also confirmed and visible from the saliency maps that these models have attributed to the inputs. Here, we observe that with a suitably trained model, Vanilla Gradient shows a minuscule degree of regularity in the saliency maps where features in all three tumor images are highlighted by the model. As with the other models, XRAI has the best interpretability for the input phenomena.

Xception shows the least visual explainability as indicated in Figure~\ref{fig:Saliency_Maps_xception}. From the input image, the Pituitary tumor located in the pituitary gland, a region below the hypothalamus is faintly attributed by all but XRAI. We can see that across all data modalities in Figures~\ref{MRI:attribution}, and~\ref{COVID:attribution}, the attribution masks give little meaningful information about the region of interest where the tumor is present although one is unsure of the COVID-19 X-ray as it is not segmented for cross-matching. Though other factors such as the dataset size, batch size, annotation quality, and data augmentation technique can considerably lead to the emergence of such characteristics, the model architecture and optimization objective have a large effect as they introduce stronger inductive priors on the space of learning functions all which we have experimentally tried to control for through hyperparameter optimization.  Moreover, this result indicates the difference between statistical correlations learned by CNNs being different from the way humans perceive and process visual stimuli.

We observed  that XRAI gives the best saliency maps as shown in the masked MRI images. We also observed segmented regions in the X-ray images with XRAI. In all the image modalities, VG and SG have coarse and partially noisy saliency maps, and can not be used to infer meaningful explanations of the model inference mechanism. The baseline choice has a major effect on the obtainable saliency map~\cite{sundararajan2017axiomatic,kapishnikov2019xrai,kapishnikov2021guided}. We used a baseline of zero pixels for all attribution methods primarily because it is information neutral. XRAI demonstrated higher interpretability compared to vanilla gradient and guided integrated gradient methods because it is more suited to deep learning-based medical image analysis tasks where the emphasis is to understand the region of interest from which a model inferred its prediction. We observed that a combination of XRAI and Blur IG can deduce feature saliency from the medical scans as 35\% of saliency maps of XRAI highlights important features that are in a close approximation of expert segmentation for the DenseNet121 model. So, utilizing multiple attribution methods can improve model interpretability for domain experts.

These results, therefore, open the possibility of not only accelerating the visual interpretability of deep neural models in medical image analysis but as well offset prepossessing such as human-in-the-loop segmentation, model debugging, and debiasing which are all crucial in real-world application use cases. The latter has an important role in low-decision risk and highly regulated domains such as healthcare. In sum, these stated use cases can rapidly advance access to needed but affordable healthcare for low-resource settings. 

However, Table~\ref{tab:Model_Architectures} in tandem with Figures ~\ref{MRI:attribution} and~\ref{COVID:attribution} show that the inductive architectural priors have to most impact on the selectivity of the receptive fields of CNNs for visual saliency analysis. CNNs perform spatial weight sharing where each filter is replicated across the entire visual field of the input~\cite{luo2016understanding}, thus, the resolution of this receptive field matters. Unlike humans, CNNs have frequency response, texture, and shape biases that are evident across all the model architectures~\cite{geirhos2018imagenet,baker2018deep}. Visual attribution methods must consider raising this notion in human-in-the-loop AI systems to ameliorate the pitfalls of the wrong attribution in deep models for real-world healthcare applications. 
\section{Conclusion}
\label{conclusion}
Deep learning models are gaining traction in ubiquitous healthcare applications from the application of vision techniques to language models. However, the inference mechanisms of these models are still an open question. In this paper, we posed the question: What do these deep learning models learn in medical images? To answer this question, we study a selection attribution framework and evaluated the framework using two widely used medical imaging modalities, namely MRI, and X-ray with  publicly available datasets. Our findings show that the robust statistical regularities learned between input-output mappings differ from biological visual stimuli processing done by humans. We show that different input attribution methods have varying degrees of explainability of the input signal. A robust representation learner and the right attribution approach are crucial to getting interpretable saliency maps of deep CNNs in medical image analysis. This is important because it will help in building human-in-the-loop computer-aided diagnostic models that not only generalize well to unseen samples but are also explainable to domain experts. Our findings indicate that deep learning models can complement the efforts of medical experts in efficiently detecting and diagnosing diseases from medical images. Thus, a human-in-the-loop approach can accelerate the adoption of neural models in medical decision-making. It provides a path toward building stakeholder trust given that healthcare requires critical evaluation of assistive technologies before adoption and general usage.

Finally, we encourage further research into volumetric medical imaging data, quantification of explainability of these visual attribution methods, developing benchmarks against which new visual attribution methods can be measured to accelerate model explainability research, and the provision of open access segmented dataset so as to test new saliency algorithms in ground truth expert segmented datasets. 
\backmatter







\section*{Declarations}
The authors declare that they have no competing interests.
\begin{itemize}
\item Funding: 
This research received no external funding.
\item Ethics approval: Not applicable
\item Consent to participate: Not applicable
\item Consent for publication: All authors have given their consent
\item Availability of data and materials:
This research used the brain tumor dataset from the School of Biomedical Engineering
Southern Medical University, Guangzhou, contains 3064 T1-weighted contrast-enhanced images with three kinds of brain tumors. The data is publicly available at \href{https://figshare.com/articles/dataset/brain_tumor_dataset/1512427}{Brain Tumor Dataset}. The Chest X-Ray dataset is publicly available at:
\href{https://www.kaggle.com/datasets/paultimothymooney/chest-xray-pneumonia}{Chest X-Ray Images (Pneumonia) Dataset}. 
\item Code availability: 
The code is available at \href{https://github.com/yusufbrima/XDNNBioimaging}{XDNNBioimaging} for reproducibility.
\item Authors' contributions: 
All the authors contributed to this work.
\end{itemize}


\bigskip

\bibliography{sn-bibliography}

\end{document}